\documentclass[10pt,twocolumn,letterpaper]{article}

\usepackage{wacv}
\usepackage{times}
\usepackage{epsfig}

\usepackage{graphics} 
\usepackage{amsmath} 
\usepackage{amssymb}  

\usepackage{color}
\usepackage{graphicx}
\usepackage{subcaption}
\usepackage{floatrow}
\definecolor{note}{rgb}{0.1,0.1,1}

\wacvfinalcopy 

\def\wacvPaperID{351} 

\ifwacvfinal\fi
\begin{document}
	
\title{Kinematically-Informed Interactive Perception: \\Robot-Generated 3D Models for Classification}

\author{Abhishek Venkataraman \\
	Robotics Institute\\ University of Michigan, Ann Arbor\\
	{\tt\small abhven@umich.edu}
	\and
	Brent Griffin \\
	Robotics Institute \&\\Electrical Engineering and Computer Science \\ University of Michigan, Ann Arbor\\
	{\tt\small griffb@umich.edu}
	\and
	Jason J. Corso \\
	Robotics Institute \&\\Electrical Engineering and Computer Science \\ University of Michigan, Ann Arbor\\
	{\tt\small jjcorso@umich.edu}
}

\maketitle
\ifwacvfinal\thispagestyle{empty}\fi

\begin{abstract}
To be useful in everyday environments, robots must be able to observe and learn about objects.
Recent datasets enable progress for classifying data into known object categories; however, it is unclear how to collect reliable object data when operating in cluttered, partially-observable environments.
In this paper, we address the problem of building complete 3D models for real-world objects using a robot platform, which can remove objects from clutter for better classification.
Furthermore, we are able to learn entirely new object categories as they are encountered, enabling the robot to classify previously unidentifiable objects during future interactions.
We build models of grasped objects using simultaneous manipulation and observation, and we guide the processing of visual data using a kinematic description of the robot to combine observations from different viewpoints and remove background noise.
To test our framework, we use a mobile manipulation robot equipped with an RGBD camera to build voxelized representations of unknown objects and then classify them into new categories.
We then have the robot remove objects from clutter to manipulate, observe, and classify them in real-time.	


\end{abstract}

\section{INTRODUCTION}
Many robots are successful in controlled environments, but robots that can adapt to our daily lives are a work in progress. 
Robots that are in our homes perform only specific tasks, such as iRobot's vacuuming Roomba. 
Recently developed hardware platforms like Fetch Robotics' Fetch \cite{wise2016fetch} and Toyota's Human Support Robot (``HSR'') \cite{UiYamaguchi2015} are trying to bridge this gap, but there are many challenges in perception that need to be addressed before personal robots are a viable product in unstructured, human environments. 

\begin{figure}[t]
	\centering
	\includegraphics[width=0.9\textwidth]{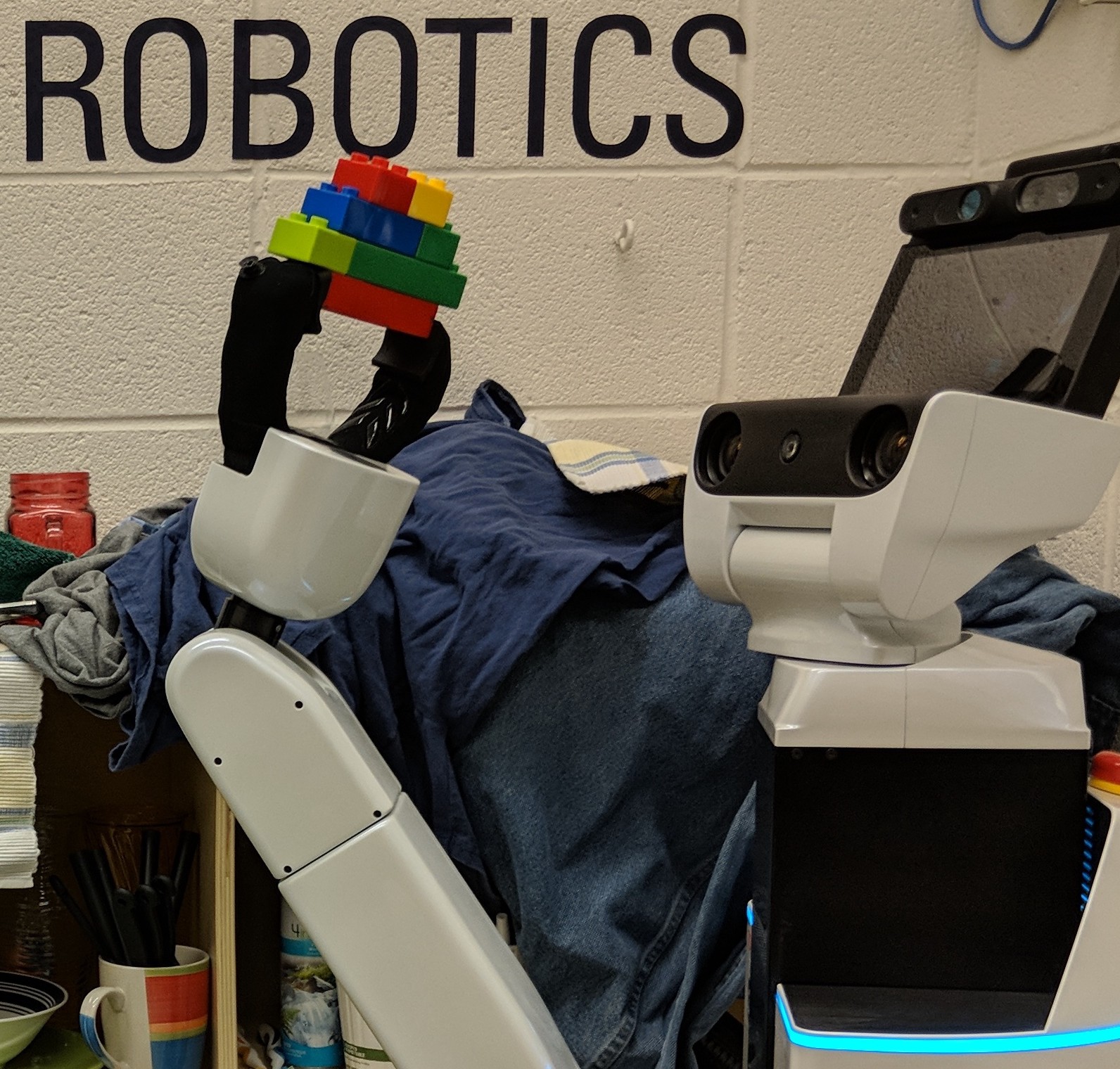}
	\caption{
		HSR manipulating and observing an unknown object away from clutter. 
		Learning-based classification methods are frequently trained using ideal instances of data. 
		To classify objects in noisy, cluttered environments, our method uses interactive perception and kinematics to isolate objects and focus observations, bridging the gap between recent advances in computer vision and robot applications. 
		\label{MICHSR}}
\end{figure}

\subsection*{Object Detection Methods}
Most home-service tasks involve interaction with household objects; naturally, a good object detector is a necessary stepping stone for the success of personal robots. 
Object detection in indoor scenes is well-studied and methods such as Mask R-CNN\cite{mask_rcnn}, Fast R-CNN \cite{girshickICCV15fastrcnn}, SSD \cite{SingleShotDetector}, and R-FCN \cite{dai16rfcn} perform well with single RGB image frames. 
The availability of large amounts of labeled training data from datasets like MS-COCO \cite{MS-COCO} and ImageNet \cite{ImageNet} are a key contributor to the success of learning-based 2D image classification methods.
One limitation of 2D image classification is not leveraging the performance benefits of 3D sensors \cite{gupta,silberman,Song_2015_CVPR}; the availability of low-cost cameras providing additional depth information (RGBD) has sparked interest in 2.5D- and 3D-based object detection.
Though not as thoroughly studied as 2D images, methods like VoxelNet \cite{voxnet} and PointNet \cite{PointNet_garcia} perform well in the 3D dataset ModelNet40  \cite{ModelNet}, which consists of CAD models for 40 objects converted to voxel grids of size $30\times30\times30.$
Unfortunately, dataset-driven classification performance can degrade in cluttered environments with occlusions or when observing objects absent during training.
Furthermore, even when application-specific datasets provide a more focused prior \cite{RGBD_object}, it is impossible to model every object that a robot will encounter in real-world environments.
Hence, it is imperative that robots learn to recognize new objects.

\subsection*{Interactive Perception}

\begin{figure*}
	\centering
	\includegraphics[width=0.975\textwidth]{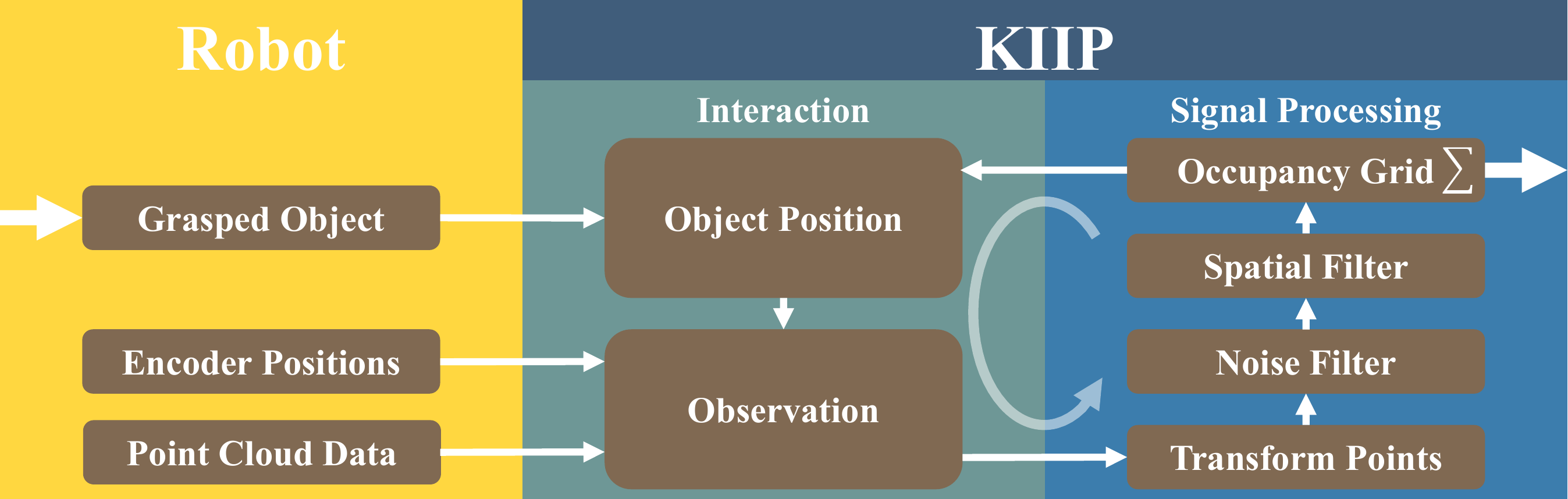}
	\caption{KIIP Process Overview. Given a robot, grasped object, and RGBD camera (yellow), KIIP performs simultaneous changes in pose and observations (green) to generate an output voxel representation (blue). The $\sum$ in Occupancy Grid indicates the summation of multiple 2.5D point cloud observations to form the final output 3D representation.}
	\label{Framework}
\end{figure*}

For robot perception, applying data-driven detection methods alone is wasting a critical asset: robots can take explicit actions to improve sensing and understanding of their environments (see Figure~\ref{MICHSR}).
Accordingly, Active Perception (AP) \cite{Bajcsy2018_active_perc, Bajcsy_active_perc} and Interactive Perception (IP) \cite{BoHaSaBrKrScSu17} are two areas of research exploiting robot-specific capabilities to improve perception.
Compared to structure from motion \cite{SFM_survey, LonguetHiggins198761}, which requires feature matching or scene flow to relate images, AP exploits knowledge of a robot's relative position to relate images and improve 3D reconstruction.
Furthermore, AP can select future view locations that improve object detection and classification \cite{Eidenberger_IROS2010,zeng2018robotic_amazon_pick_place}.
In addition to AP, IP leverages physical interactions with the environment to create novel sensory signals that otherwise would not exist.
Robots have pushed objects to improve visual segmentation \cite{ScUdAs14,ScMoAsUd13,HoKrPe14}, performed grasp-rotate-release operations  to recognize objects \cite{BeUd15}, and even shake and drop objects to learn human-provided labels through the ensuing visual, acoustic, and joint-torque signals \cite{SiScSt14}.

\subsection*{Object Classification using Interactive Perception}
In work by Krainin et al. \cite{KrHeReFo11} and Browatzki et al. \cite{BrTiMeBuWa14}, grasped objects are manipulated through many observable poses to generate a more informative collective observation.
However, rather than relying on post-processing algorithms to fuse disjoint data, in this work, RGBD-based observations are combined using only kinematic information, which is readily available using a robot's joint-encoders.
We call this process \textbf{K}inematically-\textbf{I}nformed \textbf{I}nteractive \textbf{P}erception (KIIP).
Furthermore, rather than focusing on 3D reconstruction of objects alone, we use KIIP to classify unknown objects.
To learn classifications, we use KIIP on grasped objects to rapidly train our shallow object-learning network, which can optionally work in conjunction with previous training on ModelNet40 \cite{ModelNet}.
After learning new classifications, we remove unknown objects from dense piles of clutter, observe them using KIIP, and then classify them into newly-learned object categories.
Essentially, we are using KIIP to bridge the gap between learning-based classification methods and interactive robot perception in unstructured environments.


\section{KINEMATICALLY-INFORMED \\ INTERACTIVE PERCEPTION}
\label{sec:KIIP}
This section defines the process of implementing \textbf{K}inematically-\textbf{I}nformed \textbf{I}nteractive \textbf{P}erception (KIIP).
KIIP uses simultaneous manipulation and observation of grasped objects to transform a series of 2.5D point cloud images into an accurate 3D voxel representation.
Section~\ref{sec:RA} describes the physical robot and grasped object assumptions along with sensing requirements.
Section~\ref{sec:OI} details the interactive process of object manipulation and observation.
Finally, Section~\ref{sec:SP} contains signal processing details for generating the output occupancy grid.
The overall KIIP process is depicted in Figure~\ref{Framework}.

\subsection{Robot Assumptions}
\label{sec:RA}

The primary three assumptions for implementing KIIP are that 1) a rigid object is grasped by 2) a robot equipped with sufficient degrees of freedom (DOF) and actuation to re-position the grasped target with a full rotation relative to 3) a sensor capable of generating point cloud data. 
Regarding 1), the grasped object is assumed rigid to maintain a consistent geometric shape relative to the gripper, thereby merging with the gripper as a rigid body with known, controlled changes in position.
Also, while detecting grasp positions and picking up objects is not part of this study, there are reliable methods available using similar robot configurations \cite{GuPaSaPl2016, mahler2017dex}. 
For 2), it is possible for the manipulator arm to have less dexterity, but this can result in a less complete output voxel representation. 
Example sensors for 3) include (but are not limited to) a Microsoft Kinect, ASUS Xtion, or stereo camera.

\begin{figure}
	\centering
	\includegraphics[width=0.975\textwidth]{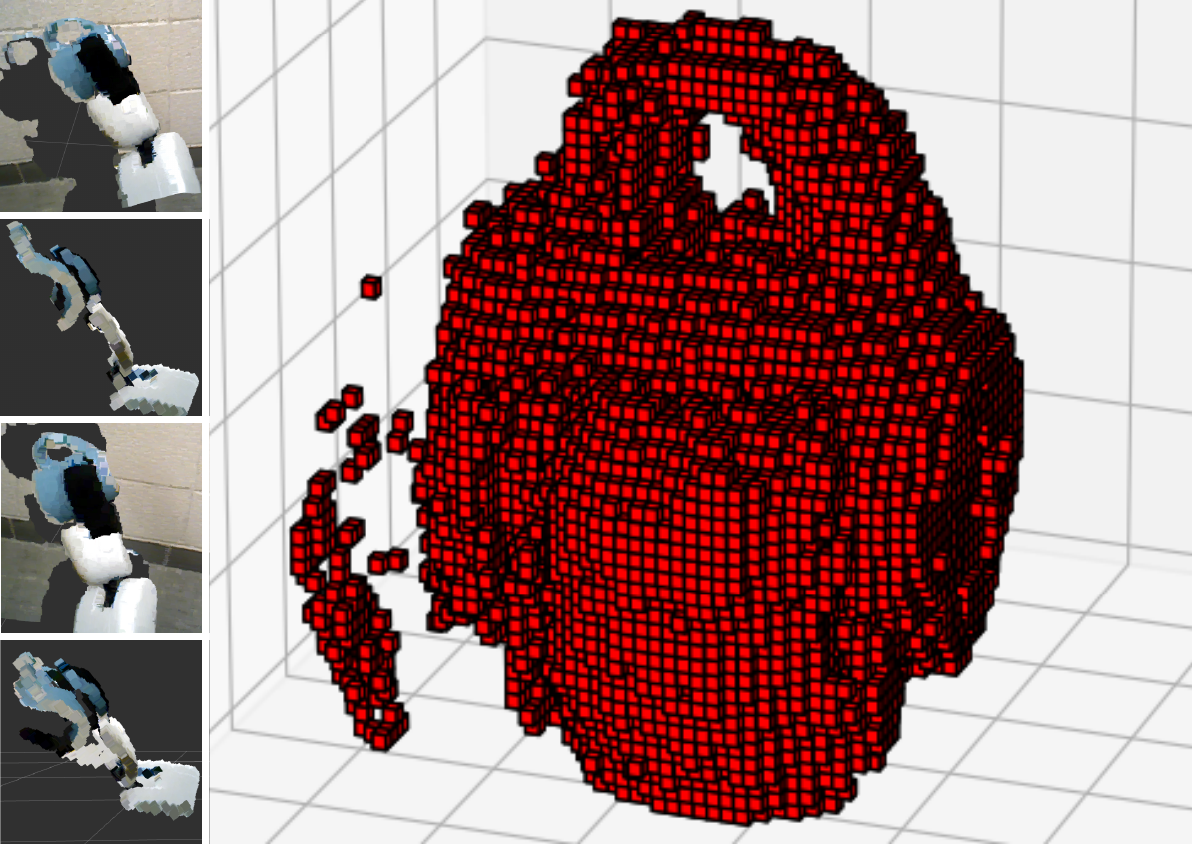}
	\caption{Point Clouds of Mug with Background and KIIP-Generated 3D Model. Point clouds are 2.5D, surface-level representations, but KIIP collects multiple 2.5D observations (left) to form the 3D voxel representation (right).\label{pc}}
\end{figure}
\begin{figure}
	\centering
	\includegraphics[width=0.975\textwidth]{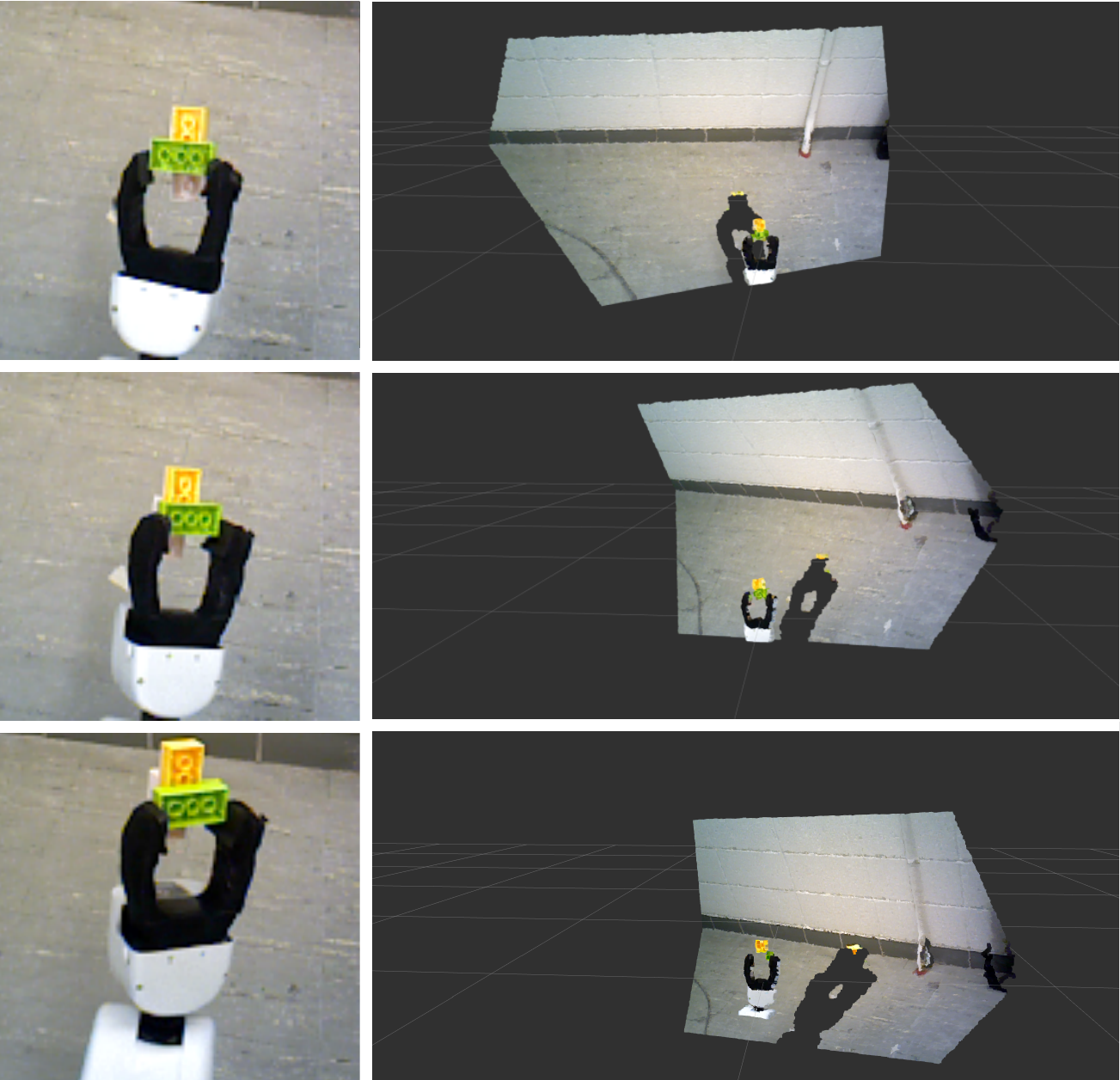}
	\caption{All point clouds are transformed from the camera frame to the gripper frame. Thus, as the arm position changes relative to the camera (left), the points associated with the grasped object stay in the same spatial location, even if the relative position of background points changes (right).}
	\label{tf_pc}
\end{figure}

In this work, we use a Toyota HSR, which has a 4-DOF manipulator arm mounted on a torso with prismatic and revolute joints and a differential drive base \cite{UiYamaguchi2015}.
For sensing, we use HSR's Xtion PRO LIVE RGBD camera, which is mounted on a 2-DOF gimbal.
The camera output is converted and published as a point cloud ROS message.


\subsection{Object Interaction and Observation}
\label{sec:OI}
The purpose of manipulating and observing the grasped object is to transform a series of 2.5D point cloud images into an accurate 3D voxel representation.
Ideally, the object is re-positioned along two separate full rotations of orthogonal axes to ensure that observations are made over all object surfaces.
Unfortunately, this level of dexterity is not feasible for most platforms.
One way to bypass this limitation is to grasp the object from multiple positions, but this eliminates the kinematic consistency between the sets of observations.

In this work, we re-position the grasped objects along 13 roll positions and then 7 pitch positions of HSR's 2-DOF wrist (20 positions total).
The 13 roll positions are linearly distributed across a $\pm$180$^\circ$ rotation of the wrist.
The first 3 pitch positions are taken at 20$^\circ$, 40$^\circ$, and 60$^\circ$ from 0$^\circ$ in the roll-wrist axis.
Next, the roll-wrist axis is rotated 180$^\circ$, where the next 4 pitch positions are taken at 0$^\circ$, 20$^\circ$, 40$^\circ$, and 60$^\circ$.
Excluding the gripper contact positions, this results in a fairly complete voxel representation, as shown in Figure~\ref{pc}.

\begin{figure*}[t!]
	\centering
	\includegraphics[width=0.98\textwidth]{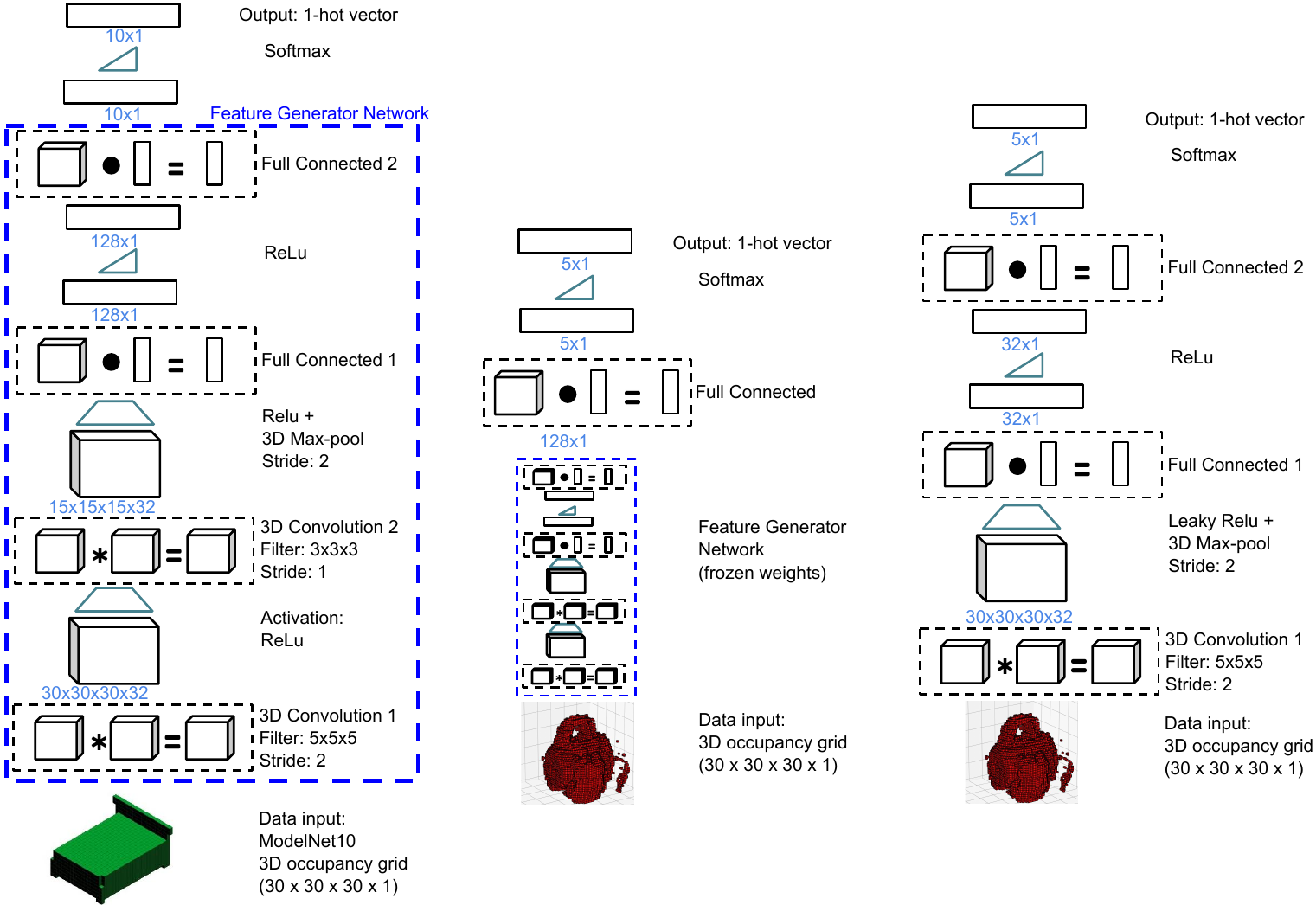}
	\caption{
Feature Generator (left), FG-OL (center), and OL-E2E (right) Network Architectures.
 \label{fig:FG}}
\end{figure*}

\subsection{Signal Processing and Output Occupancy Grid}
\label{sec:SP}

The first step of signal processing is transforming points from their native camera frame to the gripper frame.
The corresponding transformation matrix is found using the kinematic rigid-body description of the robot and joint-angle encoders.
While manually performing this transform is possible, we use the tf ROS package \cite{ROS_tf} since the point cloud messages are natively in ROS.
Transforming point cloud data to the gripper frame has two benefits.
First, it ensures that as the perspective of the grasped object changes, all of the point cloud surfaces associated with the grasped object are aligned (see Figure~\ref{tf_pc}).
Second, it enables us to apply a consistent spatial filter from within the gripper frame to remove background points and focus on the grasped object.

The second and third signal processing steps are removing signal and background noise from point cloud data.
We filter signal noise using the median value of each index point for 10 point cloud messages, which are observed while the gripper is stationary relative to the camera (more sophisticated point cloud filtering methods are also available \cite{kinect_fusion}).
In the third step, we apply our spatial filter, which removes all points outside of a 10~\rm{cm} cube extending directly from the gripper. 
This filter removes background noise and ensures that output data is focused on the grasped object.

The final signal processing step is combining all 20 point cloud perspectives into a single 3D voxel representation.
This is done by adding all of the transformed, filtered point cloud data together and projecting the sum to a $30\times30\times30$ occupancy grid.
As a final noise-reduction measure, the occupancy grid is thresholded so that all voxels corresponding to less than 4 points are 0.

\section{Three Object Classification Networks}
\label{sec:classNet}
Using the KIIP-generated $30 \times 30 \times30$ occupancy grid for grasped objects, we learn and identify object classifications using three example neural networks.
Given that the overall classification framework operates using KIIP on a physical robot, we have identified two primary goals for the architecture design of the classification networks.
First, networks should be able to train using a small number of KIIP-based instances.
Second, networks should have a rapid training time for real-time applications. 
One approach for overcoming limited instances of KIIP-collected data is pre-training on an existing 3D dataset (e.g., ModelNet), thereby enabling a network with more parameters without a large number of KIIP object interactions.
Another approach is restricting the number of layers and using a shallow network, which has the additional benefit of being rapidly trainable.
Altogether, we test three classification networks: one trained using an existing 3D dataset (FG), one trained only using KIIP data (OL-E2E), and one using both (FG-OL).

 
\subsection{Feature Generator Network (FG)}
Given the similarity of our output occupancy grid to objects in ModelNet, we chose to train a Feature Generator network (FG) on ModelNet10 \cite{ModelNet}.
For comparison to our other networks, FG acts as a dataset-trained baseline.
The FG architecture is inspired by Voxnet \cite{voxnet}, which consists of 2 conv3D layers followed by 2 fully connected layers. 
The FG network generates a 128 dimension feature vector for each of the training and testing instances (see Figure~\ref{fig:FG}, left). 
For each test instance, we assign the label of nearest (euclidean distance) training instance in the feature space. 


\subsection{FG-OL}
Our hybrid object-learning network (FG-OL) first uses the FG network to generate a 128 dimension feature vector. 
We then add a 5 neuron fully connected layer with softmax to predict the class label (see Figure~\ref{fig:FG}, center). 
We train the added network using KIIP training examples and do a forward pass of the test feature vector for predicting the label. 

\subsection{OL-E2E}
Unlike the previous network architectures, the end-to-end object-learning network (OL-E2E)  is shallow enough that it only uses KIIP data for training.
The network consists of one 3D convolution layer with 32 filters followed by 2 fully connected layers with 32 and 5 neurons respectively (see Figure~\ref{fig:FG}, right).

\section{Robot Experiments}

We design and implement object classification experiments using our KIIP framework from Section~\ref{sec:KIIP} and all three test networks from Section~\ref{sec:classNet}.
All object occupancy grids are generated using KIIP on HSR.
In the first set of experiments (Section~\ref{sec:expHouse}), we learn to classify household objects without any object-specific prior knowledge.  
As an additional challenge, because we are not limited objects conventionally found in datasets, we test our method on a novel object set of LEGO blocks assembled in various combinations (Section~\ref{sec:legoExp}).
In our experiments, we let the robot detect and pick up objects to ensure that there is significant variation in the object grasp position. 
During training, we also rotate the object so there is no inherent bias toward a single object grasp position. 
For testing, we place the object randomly within a set workspace and then let the mobile robot pick up the object. 
Discussion of experimental results is provided in Section~\ref{sec:discus}.

\subsection{Classifying Household Objects}
\label{sec:expHouse}
\begin{figure}
	\centering
	\includegraphics[width=0.975\textwidth]{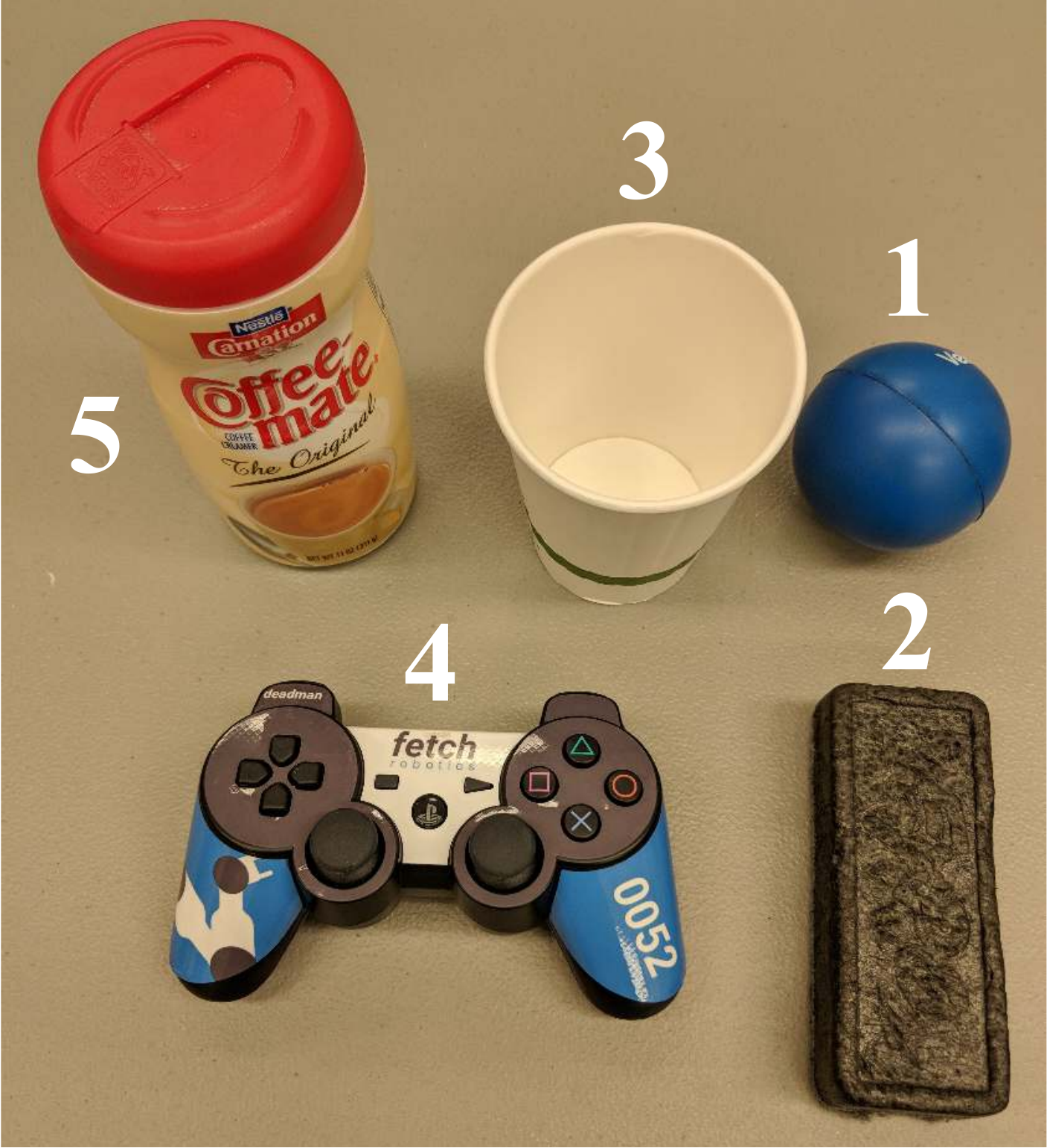}
	\caption{Household Objects for Classification. \label{fig:exp2}}
\end{figure}

The first set of experiments consists of the five common household objects shown in Figure~\ref{fig:exp2}. 
The training set consists of 8 grasp positions per object (40 instances total). 
For testing, the objects are laid out in front of HSR, with HSR picking up each object once for evaluation.
To compare the three network architectures from Section~\ref{sec:classNet} in a controlled experiment, we use the same KIIP instances for training and evaluation on all networks.
Classification results for household objects are summarized in Table~\ref{table:expRealWorld}.

\setlength{\tabcolsep}{6.25pt}
\begin{table}
	\centering
	\caption{Classification results for household objects. Figure~\ref{fig:exp2} shows the object classes with identification numbers.
		The paper cup is the only object misclassified.	
	}
	\begin{tabular}{c|c|c|c|c|c|c|c}
		\hline 
		& {Object} &\multicolumn{5}{c|}{Classification} &  \\
		& {Learning} &\multicolumn{5}{c|}{for Object} & Overall \\
		\cline{2-8}
		Network & \# Grips  & 1 & 2 & 3 & 4 & 5 & Accuracy \\
		\hline
		FG&	    8&  1&	2&	4&	4&	5&	80\% \\
		FG-OL&	8&	1&	2&	5&	4&	5&	80\% \\
		OL-E2E&	8&	1&	2&	5&	4&	5&	80\% \\
		\hline
	\end{tabular}
	\label{table:expRealWorld}
\end{table}

\begin{figure*}
	\centering
	\includegraphics[width=0.67\textwidth]{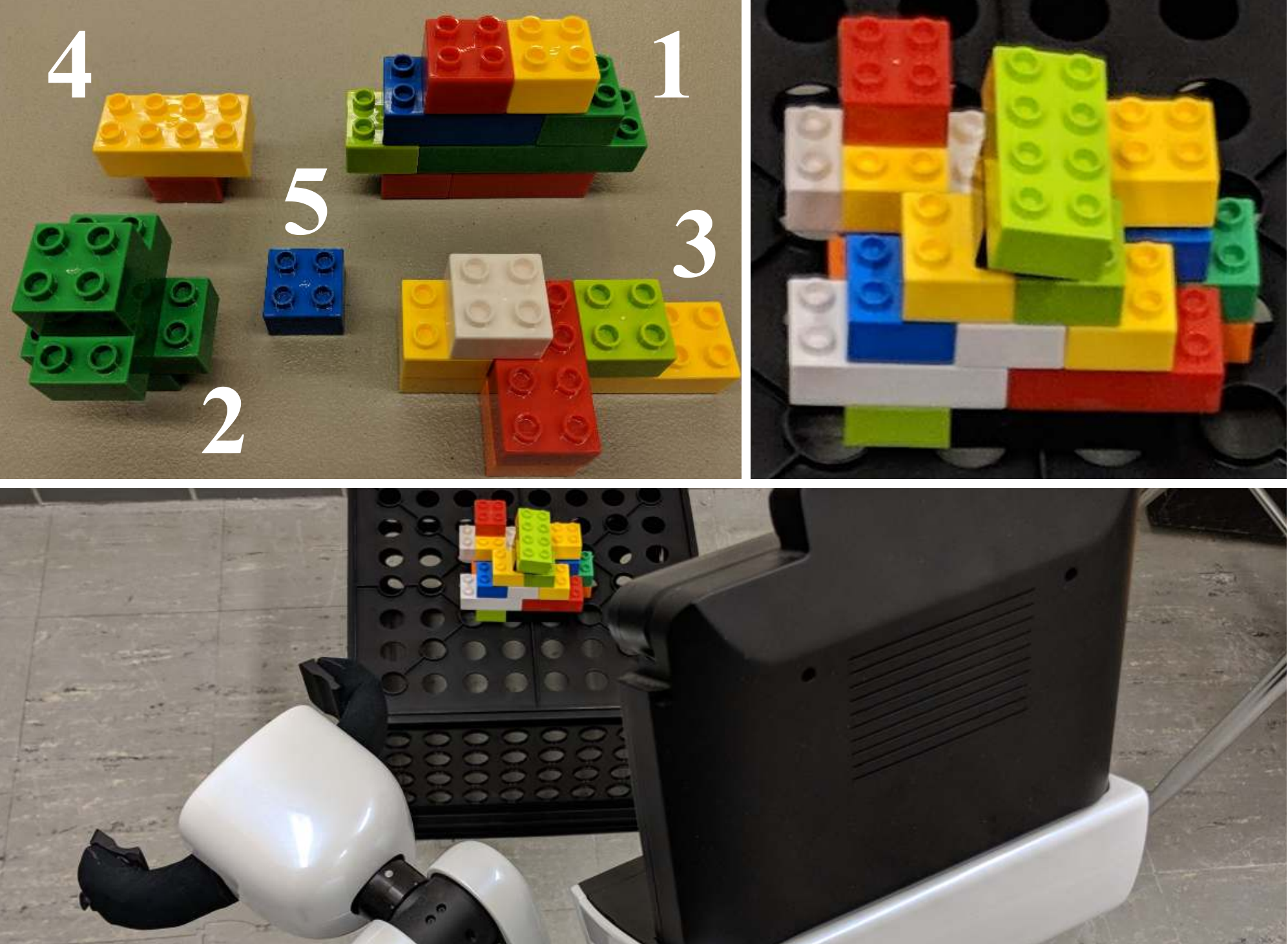}
	\caption{Unconventional Objects for Classification. As an additional challenge, we use unknown LEGO objects to build KIIP-models and train our classification networks (top left). Then, using additional instances of the unknown objects (top right), HSR identifies and grasps the top-most object in the pile to classify it in the appropriate newly-learned category (bottom). 
	 The LEGO objects have mainly flat surfaces and less contour variability compared to the set of household objects.
		\label{fig:exp1}}
\end{figure*}


\setlength{\tabcolsep}{4pt}
\begin{table}
	\centering
	\caption{Classification results for LEGO objects. Figure~\ref{fig:exp1} shows the object classes with identification numbers. Bold text indicates the best performer for each training dataset.}
	\begin{tabular}{c|c|c|c|c|c|c|c|c}
		\hline 
		& \multicolumn{2}{c|}{Object} &\multicolumn{5}{c|}{Classification} &  \\
		& \multicolumn{2}{c|}{Learning} &\multicolumn{5}{c|}{for Object} & Overall \\
		\cline{2-8}
		Network & \# Grips &  Time (\rm{s})  & 1 & 2 & 3 & 4 & 5 & Accuracy \\
		\hline
		FG & 1 & NA & 3 & 2 & 1 & 4 & 4 & 40\% \\
		FG-OL & 1	& 22.5	& 1 &	3	& 3&	2&	5&	\bf{60\%} \\
		OL-E2E & 1	&\bf{14}	&2	&2	&2	&4	&5	&\bf{60\%} \\
		\hline
		FG&	2&	NA&	3&	2&	3&	5&	5&	40\% \\
		FG-OL&	2&	21.5&	3&	2&	3&	2&	5&	40\% \\
		OL-E2E&2	&\bf{15}	&3	&2	&2	&4	&5	&\bf{60\%} \\
		\hline
		FG&	4&	NA&	2&	1&	3&	1&	4&	20\% \\
		FG-OL	&4	&20.5&	2&	2&	2&	4&	5&	\bf{60\%} \\
		OL-E2E&	4&	\bf{14}&	2&	2&	2&	5&	5&	40\% \\
		\hline
		FG&	8&	NA&	2&	4&	2&	3&	4&	20\% \\
		FG-OL&	8&	43.5&	1&	2&	1&	2&	4&	40\% \\
		OL-E2E&	8&	\bf{16.5}&	3&	2&	2&	4&	5&	\bf{60\%} \\
		\hline
		FG&16&	NA&	2&	4&	2&	1&	4&	0\% \\
		FG-OL	&16&	62.5&	1&	2&	1&	3&	5&	\bf{60\%} \\
		OL-E2E	&16&	\bf{29.5}&	3&	3&	3&	4&	5&	\bf{60\%} \\	 
		\hline
	\end{tabular}
	\label{table:expLego}
\end{table}

\subsection{Classifying Unconventional Objects}
\label{sec:legoExp}

To test our learning and classification framework on unconventional objects not found in datasets, we perform a second set of experiments using the five unique LEGO structures shown in Figure~\ref{fig:exp1}.
We first collect KIIP data instances with 16 grip positions per object (80 instances total).
To evaluate changes in classification performance against the number of unique KIIP object grasps used for training, we evaluate all three networks using multiple subsets of the total training set. 
To form the evaluation set of objects, LEGO structures from the training set are re-built using new blocks (changing object appearance, but not structure). 
To demonstrate our approach of removing objects from cluttered, partially-observable environments, the evaluation experiments begins with the LEGO structures densely piled together.
To ensure that we have random grip positions, we let HSR grab the topmost object in the pile, use KIIP to classify the object, and then discard it; we repeat this process until all of the objects in the pile are removed and classified.
As in Section~\ref{sec:expHouse}, we use the same KIIP instances for training and evaluation on all networks.
Figure~\ref{fig:exp1} shows the objects used for training and the pile of unknown objects used for evaluation.
Classification results of the experiment are summarized in Table~\ref{table:expLego}.

\subsection{Discussion of Experimental Results}
\label{sec:discus}

For the household object experiments, all of the classification networks achieved an 80\% classification accuracy.
We see these experiments as evidence supporting the practicality of robots generating KIIP-based 3D models for object learning and classification.
Furthermore, we are encouraged that the FG network, which is only trained on dataset object models, is able to classify household objects using KIIP-based object interactions.

For the more challenging LEGO object experiments, we find that the two KIIP-object trained classification networks perform best, with the OL-E2E network having an edge in overall accuracy across the various training configurations.
As an additional consideration, the shallow OL-E2E network requires no \textit{a-priori} object knowledge and has the fastest training time across all configurations (as fast as 14 seconds on single-GPU hardware).
We take the LEGO object experiments to be an indicator that, when encountering new, unconventional objects, it is practical and beneficial to learn new categories of classification using KIIP-based object interactions for training. 

Finally, we emphasize that the three networks used in this work are experimental.
Developing new architectures or incorporating object appearance (e.g., color, texture, and product labels) should dramatically improve classification performance.

\section{Conclusions}

In this work, we introduce a framework for generating 3D representations of objects from a manipulation robot equipped with point cloud sensing.
Compared to passive perception, our approach leverages actuation of the robot through object interactions, and our kinematically-informed interactive perception makes use of joint encoders that would have been otherwise wasted.
For our implementation on a mobile robot with depth sensing, we were able to achieve an 80\% classification accuracy on household objects using a network trained only on data collected by the robot itself.
We find that our approach to interactive perception is a useful tool for implementing recent learning-based classification methods on a robot operating and collecting data in real-world environments.

For reproducibility, we will provide source code with the final paper.

\newpage

{\small
	\bibliographystyle{ieee}
	\bibliography{KIIP}

\begin{thebibliography}{10}\itemsep=-1pt

\bibitem{SFM_survey}
J.~K. Aggarwal and N.~Nandhakumar.
\newblock On the computation of motion from sequences of images-a review.
\newblock {\em Proceedings of the IEEE}, 76(8):917--935, Aug 1988.

\bibitem{Bajcsy_active_perc}
R.~Bajcsy.
\newblock Active perception.
\newblock {\em Proceedings of the IEEE}, 76(8):966--1005, Aug 1988.

\bibitem{Bajcsy2018_active_perc}
R.~Bajcsy, Y.~Aloimonos, and J.~K. Tsotsos.
\newblock Revisiting active perception.
\newblock {\em Autonomous Robots}, 42(2):177--196, Feb 2018.

\bibitem{BeUd15}
R.~Bevec and A.~Ude.
\newblock Pushing and grasping for autonomous learning of object models with
  foveated vision.
\newblock In {\em 2015 International Conference on Advanced Robotics (ICAR)},
  pages 237--243, July 2015.

\bibitem{BoHaSaBrKrScSu17}
J.~Bohg, K.~Hausman, B.~Sankaran, O.~Brock, D.~Kragic, S.~Schaal, and G.~S.
  Sukhatme.
\newblock Interactive perception: Leveraging action in perception and
  perception in action.
\newblock {\em IEEE Transactions on Robotics}, 33(6):1273--1291, Dec 2017.

\bibitem{BrTiMeBuWa14}
B.~Browatzki, V.~Tikhanoff, G.~Metta, H.~H. B?lthoff, and C.~Wallraven.
\newblock Active in-hand object recognition on a humanoid robot.
\newblock {\em IEEE Transactions on Robotics}, 30(5):1260--1269, Oct 2014.

\bibitem{Eidenberger_IROS2010}
R.~Eidenberger and J.~Scharinger.
\newblock Active perception and scene modeling by planning with probabilistic
  6d object poses.
\newblock In {\em 2010 IEEE/RSJ International Conference on Intelligent Robots
  and Systems}, pages 1036--1043, Oct 2010.

\bibitem{ROS_tf}
T.~Foote.
\newblock tf: The transform library.
\newblock In {\em Technologies for Practical Robot Applications (TePRA), 2013
  IEEE International Conference on}, Open-Source Software workshop, pages 1--6,
  April 2013.

\bibitem{PointNet_garcia}
A.~Garcia-Garcia, F.~Gomez-Donoso, J.~Garcia-Rodriguez, S.~Orts-Escolano,
  M.~Cazorla, and J.~Azorin-Lopez.
\newblock Pointnet: A 3d convolutional neural network for real-time object
  class recognition.
\newblock In {\em 2016 International Joint Conference on Neural Networks
  (IJCNN)}, pages 1578--1584, July 2016.

\bibitem{girshickICCV15fastrcnn}
R.~Girshick.
\newblock Fast r-cnn.
\newblock In {\em International Conference on Computer Vision ({ICCV})}, 2015.

\bibitem{GuPaSaPl2016}
M.~Gualtieri, A.~ten Pas, K.~Saenko, and R.~Platt.
\newblock High precision grasp pose detection in dense clutter.
\newblock In {\em 2016 IEEE/RSJ International Conference on Intelligent Robots
  and Systems (IROS)}, pages 598--605, Oct 2016.

\bibitem{gupta}
S.~Gupta, R.~Girshick, P.~Arbel{\'a}ez, and J.~Malik.
\newblock Learning rich features from rgb-d images for object detection and
  segmentation.
\newblock In D.~Fleet, T.~Pajdla, B.~Schiele, and T.~Tuytelaars, editors, {\em
  Computer Vision -- ECCV 2014}, pages 345--360, Cham, 2014. Springer
  International Publishing.

\bibitem{mask_rcnn}
K.~He, G.~Gkioxari, P.~Dollár, and R.~Girshick.
\newblock Mask r-cnn.
\newblock In {\em 2017 IEEE International Conference on Computer Vision
  (ICCV)}, pages 2980--2988, Oct 2017.

\bibitem{dai16rfcn}
K.~H. J.~S. Jifeng~Dai, Yi~Li.
\newblock {R-FCN}: Object detection via region-based fully convolutional
  networks.
\newblock {\em arXiv preprint arXiv:1605.06409}, 2016.

\bibitem{KrHeReFo11}
M.~Krainin, P.~Henry, X.~Ren, and D.~Fox.
\newblock Manipulator and object tracking for in-hand 3d object modeling.
\newblock {\em The International Journal of Robotics Research},
  30(11):1311--1327, 2011.

\bibitem{RGBD_object}
K.~Lai, L.~Bo, X.~Ren, and D.~Fox.
\newblock A large-scale hierarchical multi-view rgb-d object dataset.
\newblock In {\em 2011 IEEE International Conference on Robotics and
  Automation}, pages 1817--1824, May 2011.

\bibitem{MS-COCO}
T.-Y. Lin, M.~Maire, S.~Belongie, J.~Hays, P.~Perona, D.~Ramanan,
  P.~Doll{\'a}r, and C.~L. Zitnick.
\newblock Microsoft coco: Common objects in context.
\newblock In D.~Fleet, T.~Pajdla, B.~Schiele, and T.~Tuytelaars, editors, {\em
  Computer Vision -- ECCV 2014}, pages 740--755, Cham, 2014. Springer
  International Publishing.

\bibitem{SingleShotDetector}
W.~Liu, D.~Anguelov, D.~Erhan, C.~Szegedy, S.~Reed, C.-Y. Fu, and A.~C. Berg.
\newblock Ssd: Single shot multibox detector.
\newblock In B.~Leibe, J.~Matas, N.~Sebe, and M.~Welling, editors, {\em
  Computer Vision -- ECCV 2016}, pages 21--37, Cham, 2016. Springer
  International Publishing.

\bibitem{LonguetHiggins198761}
H.~Longuet-Higgins.
\newblock A computer algorithm for reconstructing a scene from two projections.
\newblock In M.~A. Fischler, , and O.~Firschein, editors, {\em Readings in
  Computer Vision}, pages 61 -- 62. Morgan Kaufmann, San Francisco (CA), 1987.

\bibitem{mahler2017dex}
J.~Mahler, J.~Liang, S.~Niyaz, M.~Laskey, R.~Doan, X.~Liu, J.~A. Ojea, and
  K.~Goldberg.
\newblock Dex-net 2.0: Deep learning to plan robust grasps with synthetic point
  clouds and analytic grasp metrics.
\newblock 2017.

\bibitem{voxnet}
D.~Maturana and S.~Scherer.
\newblock Voxnet: A 3d convolutional neural network for real-time object
  recognition, 09 2015.

\bibitem{kinect_fusion}
R.~A. Newcombe, S.~Izadi, O.~Hilliges, D.~Molyneaux, D.~Kim, A.~J. Davison,
  P.~Kohi, J.~Shotton, S.~Hodges, and A.~Fitzgibbon.
\newblock Kinectfusion: Real-time dense surface mapping and tracking.
\newblock In {\em 2011 10th IEEE International Symposium on Mixed and Augmented
  Reality}, pages 127--136, Oct 2011.

\bibitem{ImageNet}
O.~Russakovsky, J.~Deng, H.~Su, J.~Krause, S.~Satheesh, S.~Ma, Z.~Huang,
  A.~Karpathy, A.~Khosla, M.~Bernstein, A.~C. Berg, and L.~Fei-Fei.
\newblock {ImageNet Large Scale Visual Recognition Challenge}.
\newblock {\em International Journal of Computer Vision (IJCV)},
  115(3):211--252, 2015.

\bibitem{ScMoAsUd13}
D.~Schiebener, J.~Morimoto, T.~Asfour, and A.~Ude.
\newblock Integrating visual perception and manipulation for autonomous
  learning of object representations.
\newblock {\em Adaptive Behavior}, 21(5):328--345, 2013.

\bibitem{ScUdAs14}
D.~Schiebener, A.~Ude, and T.~Asfour.
\newblock Physical interaction for segmentation of unknown textured and
  non-textured rigid objects.
\newblock In {\em 2014 IEEE International Conference on Robotics and Automation
  (ICRA)}, pages 4959--4966, May 2014.

\bibitem{silberman}
N.~Silberman, D.~Hoiem, P.~Kohli, and R.~Fergus.
\newblock Indoor segmentation and support inference from rgbd images.
\newblock In A.~Fitzgibbon, S.~Lazebnik, P.~Perona, Y.~Sato, and C.~Schmid,
  editors, {\em Computer Vision -- ECCV 2012}, pages 746--760, Berlin,
  Heidelberg, 2012. Springer Berlin Heidelberg.

\bibitem{SiScSt14}
J.~Sinapov, C.~Schenck, and A.~Stoytchev.
\newblock Learning relational object categories using behavioral exploration
  and multimodal perception.
\newblock In {\em 2014 IEEE International Conference on Robotics and Automation
  (ICRA)}, pages 5691--5698, May 2014.

\bibitem{Song_2015_CVPR}
S.~Song, S.~P. Lichtenberg, and J.~Xiao.
\newblock Sun rgb-d: A rgb-d scene understanding benchmark suite.
\newblock In {\em The IEEE Conference on Computer Vision and Pattern
  Recognition (CVPR)}, June 2015.

\bibitem{HoKrPe14}
H.~van Hoof, O.~Kroemer, and J.~Peters.
\newblock Probabilistic segmentation and targeted exploration of objects in
  cluttered environments.
\newblock {\em IEEE Transactions on Robotics}, 30(5):1198--1209, Oct 2014.

\bibitem{wise2016fetch}
M.~Wise, M.~Ferguson, D.~King, E.~Diehr, and D.~Dymesich.
\newblock Fetch and freight: Standard platforms for service robot applications.
\newblock In {\em Workshop on Autonomous Mobile Service Robots}, 2016.

\bibitem{ModelNet}
Z.~Wu, S.~Song, A.~Khosla, F.~Yu, L.~Zhang, X.~Tang, and J.~Xiao.
\newblock 3d shapenets: A deep representation for volumetric shapes.
\newblock In {\em 2015 IEEE Conference on Computer Vision and Pattern
  Recognition (CVPR)}, pages 1912--1920, June 2015.

\bibitem{UiYamaguchi2015}
U.~Yamaguchi, F.~Saito, K.~Ikeda, and T.~Yamamoto.
\newblock Hsr, human support robot as research and development platform.
\newblock {\em The Abstracts of the international conference on advanced
  mechatronics : toward evolutionary fusion of IT and mechatronics : ICAM},
  2015.6:39--40, 2015.

\bibitem{zeng2018robotic_amazon_pick_place}
A.~Zeng, S.~Song, K.-T. Yu, E.~Donlon, F.~R. Hogan, M.~Bauza, D.~Ma, O.~Taylor,
  M.~Liu, E.~Romo, N.~Fazeli, F.~Alet, N.~C. Dafle, R.~Holladay, I.~Morona,
  P.~Q. Nair, D.~Green, I.~Taylor, W.~Liu, T.~Funkhouser, and A.~Rodriguez.
\newblock Robotic pick-and-place of novel objects in clutter with
  multi-affordance grasping and cross-domain image matching.
\newblock In {\em Proceedings of the IEEE International Conference on Robotics
  and Automation}, 2018.

\end{thebibliography}


\begin{thebibliography}{1}\itemsep=-1pt

\bibitem{Alpher02}
A.~Alpher.
\newblock Frobnication.
\newblock {\em Journal of Foo}, 12(1):234--778, 2002.

\bibitem{Alpher03}
A.~Alpher and J.~P.~N. Fotheringham-Smythe.
\newblock Frobnication revisited.
\newblock {\em Journal of Foo}, 13(1):234--778, 2003.

\bibitem{Alpher04}
A.~Alpher, J.~P.~N. Fotheringham-Smythe, and G.~Gamow.
\newblock Can a machine frobnicate?
\newblock {\em Journal of Foo}, 14(1):234--778, 2004.

\bibitem{Authors06b}
Authors.
\newblock Frobnication tutorial, 2006.
\newblock Supplied as additional material {\tt tr.pdf}.

\bibitem{Authors06}
Authors.
\newblock The frobnicatable foo filter, 2011.
\newblock Face and Gesture submission ID 324. Supplied as additional material
  {\tt fg324.pdf}.

\end{thebibliography}
}

\end{document}